\documentclass[11pt]{article}

\usepackage[final]{EMNLP2022}

\usepackage{times}
\usepackage{latexsym}

\usepackage[T1]{fontenc}

\usepackage[utf8]{inputenc}

\usepackage{microtype}

\usepackage{inconsolata}

\usepackage[dvips]{graphicx}
\usepackage{multirow}
\usepackage{diagbox}
\usepackage{arydshln}
\usepackage{algorithm,algorithmicx}
\usepackage{algpseudocode}
\usepackage{booktabs}

\title{Extending the Subwording Model of Multilingual Pretrained Models
  for New Languages}

\author{Kenji Imamura \and Eiichiro Sumita \\
National Institute of Information and Communications Technology \\
3-5 Hikaridai, Seika-cho, Soraku-gun, Kyoto 619-0289, Japan \\
\texttt{\{kenji.imamura,eiichiro.sumita\}@nict.go.jp}}

\begin{document}
\maketitle

\begin{abstract}

Multilingual pretrained models are effective for machine translation
and cross-lingual processing because they contain multiple languages
in one model.  However, they are pretrained after their tokenizers are
fixed; therefore it is difficult to change the vocabulary after
pretraining.  When we extend the pretrained models to new languages,
we must modify the tokenizers simultaneously.

In this paper, we add new subwords to the SentencePiece tokenizer to
apply a multilingual pretrained model to new languages (Inuktitut in
this paper).
In our experiments, we segmented Inuktitut sentences into subwords
without changing the segmentation of already pretrained languages, and
applied the mBART-50 pretrained model to English-Inuktitut
translation.

\end{abstract}

\section{Introduction}
\label{sec-introduction}

Various pretrained models have been released in recent years.  Among
them, multilingual pretrained models, for example, multilingual BERT
(mBERT) \cite{devlin-etal-2019-bert}, XLM-RoBERTa
\cite{conneau2020unsupervised}, mBART
\cite{liu-etal-2020-multilingual-denoising,lewis-etal-2020-bart}, and mT5
\cite{xue-etal-2021-mt5}, are effective for cross-lingual processing
and machine translation because they contain multiple languages in one
model.  However, the vocabulary sets of multilingual pretrained models
are fixed before pretraining, and it is difficult to change the
vocabulary after pretraining.  This restriction becomes a problem when
we enhance the multilingual pretrained models for new languages.

The vocabulary set can be added to the pretrained models by extending
word embedding tables \cite{wang-etal-2020-extending}.  However, these
models are pretrained after the tokenizers (and their models) are
fixed.  When we change the vocabulary set in the pretrained models, we
must simultaneously modify the tokenizers.

If a new language uses known letters, an existing multilingual
tokenizer can segment input sentences into known subwords, which are
included in the vocabulary of the multilingual pretrained models.
Therefore, tokenization is not a crucial problem in this case
(even though it is not optimal).
For instance, mBART-50 \cite{tang2020multilingual}, which is a
multilingual encoder-decoder pretrained model, supports 52 languages
and uses SentencePiece
\cite{kudo-richardson-2018-sentencepiece,kudo-2018-subword}%
\footnote{\url{https://github.com/google/sentencepiece}} as the
tokenizer.  The model of SentencePiece for mBART-50 was learned using
a corpus that consisted of 100 languages, which included all languages
of the pretrained model.  When we add a new language to mBART-50, we
can divert the tokenizing model without modification if the letters of
the language are already included in the tokenizer.

If the letters of new languages are not supported by the tokenizer,
one solution is to transliterate unseen letters into Latin letters
\cite{muller-etal-2021-unseen}.  This approach is appropriate for
encoder models such as mBERT.  However, it is not suitable for
decoders because we should generate the unique letters of the
languages.

Another solution is to separate the tokenizer of the new language from
the others.  However, we want to enhance the tokenizer model while
maintaining the vocabulary set of the original languages because
sentences in the new language often contain words in the original
languages (e.g., numbers, named entities, and code switching).

In this paper, we focus on the tokenizers of multilingual pretrained
models.  Specifically, we add new subwords to the SentencePiece
tokenizer to apply an mBART-50 model to new languages.  The task in
this paper is English-Inuktitut translation, which was a shared task
at the WMT-20 conference \cite{wmt-2020-machine}.  Because the
Inuktitut language is not supported in either the mBART-50 model or
its SentencePiece tokenizer, we add Inuktitut subwords to the models.
To add new subwords to the SentencePiece model, we must not only add
new entries but also estimate their costs.  In this paper, we estimate
the costs of new subwords and tokenize Inuktitut text into subwords
without changing the original languages.

The remainder of this paper is organized as follows: In Section
\ref{sec-related-work}, we explain related work, which includes
studies on the adaptation of multilingual pretrained models to new
languages (Section \ref{sec-additional-language}) and an overview of
the SentencePiece tokenizer (Section \ref{sec-sentencepiece}).  In
Section \ref{sec-subword-addition}, we describe our proposal, that is,
the addition of subwords to the SentencePiece model.  In Section
\ref{sec-experiments}, we evaluate the tokenization and translation
results of English-Inuktitut translation, and we conclude the paper in
Section \ref{sec-conclusions}.

\section{Related Work}
\label{sec-related-work}

\subsection{Adaptation of Multilingual Pretrained Models to New Languages}
\label{sec-additional-language}

In this section, we mainly discuss the tokenizers required to add new
languages to multilingual pretrained models.

\newcite{ebrahimi-kann-2021-adapt} enhanced the multilingual
pretrained model XLM-R to 1,600 languages.  They trained a new
SentencePiece model from scratch using the multilingual corpora of all
languages (i.e., original and new languages) and did not use the
tokenizing model of XLM-R.  Therefore, the training corpora of the
original languages are necessary.

\newcite{muller-etal-2021-unseen} added new languages to the mBERT
models.  If the new language consisted of unseen letters, they
transliterated them to Latin characters and applied the WordPiece
tokenizer \cite{WordPiece2012}.  Therefore, the original tokenization
model was diverted.  Although this method is appropriate for encoder
models, we cannot generate language-specific letters using decoder
models.

\newcite{wang-etal-2020-extending} added new languages to mBERT models
by extending their word embedding tables.  Although the authors did not
describe the details of the tokenizers, we assume that unknown letters
were segmented into letters because mBERT uses the WordPiece tokenizer.

\newcite{artetxe-etal-2020-cross} trained a tokenizer independently
for new languages.  However, the vocabulary of the new tokenizer was
mismatched with the subwords of the original languages because new
languages often contain words in the original languages (e.g.,
numbers, named entities, and code switching).
Similarly, \newcite{arxiv.2012.15562} enhanced mBERT and XLM-R
using tokenizers trained separately for each language.

Our objective is to provide subword tokenization to a new language
while maintaining the tokenization of the original languages, and
apply multilingual pretrained models to the new language.  By
maintaining the tokenization of the original languages, we leverage
the effects of the pretrained models.

\subsection{SentencePiece Tokenizer}
\label{sec-sentencepiece}

SentencePiece is a tokenizer that directly tokenizes input text into
subwords (called lossless encoding).  It is used as the tokenizer for
multilingual pretrained models because it can process languages in
which word boundaries are explicitly indicated by a space character
(e.g., English) and those without a word boundary (e.g., Chinese and
Japanese) in the same manner.

Although SentencePiece supports byte-pair encoding
\cite{sennrich-etal-2016-neural}, we discuss the unigram model, which
is the default model of SentencePiece that is used for mBART-50.

\subsubsection{Subword Tokenization of SentencePiece}
\label{sec-analysis}

\begin{figure*}
\begin{center}
\includegraphics[clip,width=140mm]{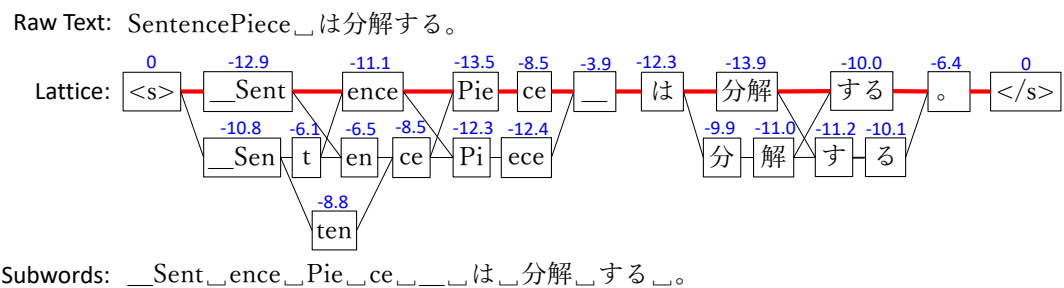}
\caption{Example of subword tokenization using the SentencePiece
  unigram model.  The '$\sqcup$' symbol indicates the space 
  (U+0020).  The values above subwords indicate the log-likelihood of the
  unigrams.  The red line indicates the Viterbi path.}
\label{fig-lattice}
\end{center}
\end{figure*}

To achieve lossless encoding, SentencePiece converts input text into
text without space characters by substituting the spaces for a special
letter (Unicode letter U+2581 by default).  Then it tokenizes
converted text into subwords as follows (Figure \ref{fig-lattice})
\cite{manning-fsnlp1999}.  Note that this is identical to the
morphological analysis method for unsegmented languages (e.g., MeCab
\cite{kudo-etal-2004-applying}).

\begin{enumerate}
\item
The input text is matched with the unigram model (corresponding to a
morphological analysis dictionary) of the tokenizer, and all
subword candidates obtained from the model are structured into a
lattice.

\item
The Viterbi search is applied to the lattice to search for the best
path (i.e., the path of highest likelihood).  A subword
sequence is output on the Viterbi path.
\end{enumerate}

\subsubsection{Learning the Unigram Model}
\label{sec-model-training}

The unsupervised learning algorithm of hidden Markov models
\cite{manning-fsnlp1999} is applied to the learning of the unigram
model.  The procedure is as follows.  The algorithm simply eliminates
subwords in the initial model, where the vocabulary of the final model
becomes a subset of that of the initial model.

\begin{enumerate}
\item Build the initial model from the training corpora.

  \begin{enumerate}\itemsep=0mm
  \item Obtain subword candidates by acquiring substrings in the
    corpora using a suffix array.

  \item Supply each subword candidate with
    a likelihood computed from the relative frequency in the corpora.
  \end{enumerate}

\item Iterate the following procedure until the vocabulary size of the
  model becomes predefined.

  \begin{enumerate}\itemsep=0mm
  \item Update the likelihood of the subwords (twice) using the EM
    algorithm.

    \begin{enumerate}\itemsep=0mm
    \item \textbf{E Step:}\\
      Analyze sentences in the corpora using the current model, and
      compute the likelihood of each subword in the sentence by
      applying the forward-backward algorithm.

    \item \textbf{M Step:}\\
      Collect the likelihoods of all subwords in the corpora and
      update the model.
    \end{enumerate}

  \item Eliminate the low-likelihood subwords from the model (e.g.,
    20 percent of all subwords).
  \end{enumerate}
\end{enumerate}

To balance speed and quallity, the step 2.(a) is performed only twice
because we do not need fully accurate likelihoods in early iterations
(it is enough to specify subword candidates for elimination).  By
iterating the step 2., we can obtain accurate likelihoods close to
convergence.

\section{Adding Subwords to the Unigram Model}
\label{sec-subword-addition}

If the new language uses known letters, an original multilingual
tokenizer can segment input sentences into known subwords, which are
included in the vocabulary of the multilingual pretrained models.
Therefore, tokenization is not a crucial problem in this case
even though it is not optimal.

By contrast, when the new language uses unknown letters, we cannot
determine the vocabulary of the pretrained model because the string of
an entire sentence becomes unknown words in lossless encoding.  A way
to solve this problem is to segment the string into letters and
recognize them as tokens.  However, this method is disadvantageous to
downstream tasks such as machine translation because token sequences
become very long.  Our method segments sentences in the new language
into subwords while restricting both the lengths of token sequences
and vocabulary size.

In this paper, we add only subwords that include unknown letters to
the tokenizer to segment the new language into subwords without changing
the results of existing corpora.

\begin{algorithm}[t]
\caption{Additional Subword Model Learning}
\label{algo-additional-model}
{\small
\begin{algorithmic}[1]
  \renewcommand{\algorithmicrequire}{\textbf{Input:}}
  \renewcommand{\algorithmicensure}{\textbf{Output:}}
  \Require Corpus $C$, vocabulary size $V$, original model $M_{org}$
  \Ensure  Additional model $M$
  \State $M \leftarrow \textsc{GenerateInitModel}(C)$ 
  \label{line-initialize}
  \While{$|M| > V$}
  \label{line-main-loop-start}
  \State $M \leftarrow \textsc{UpdateLikelihood}(M, M_{org}, C)$
  \State $M \leftarrow \textsc{UpdateLikelihood}(M, M_{org}, C)$
  \State \parbox[t]{\dimexpr\linewidth-\leftmargin}{%
    Eliminate low-likelihood subwords $\langle sw, l \rangle$ from $M$.
  }
  \EndWhile
  \label{line-main-loop-end}
  \State\Return $M$ 
  
  \Statex
  \Function{GenerateInitModel}{$C$}
  \label{line-init-model-start}
  \State $M \leftarrow \emptyset$
  \ForAll{substring $sw \in C$}
  \State \parbox[t]{\dimexpr\linewidth-\leftmargin-\labelwidth}{%
    Compute the likelihood $l$ of $sw$ from the relative frequency in $C$.
  }
  \State $M \leftarrow M \cup \langle sw, l \rangle$
  \EndFor
  \State\Return $M$
  \EndFunction
  \label{line-init-model-end}

  \Statex
  \Function{UpdateLikelihood}{$M, M_{org}, C$}
  \label{line-analyze-corpus-start}
  \ForAll{sentence $S \in C$}
  \Comment{E step}
  \State \parbox[t]{\dimexpr\linewidth-\leftmargin-\labelwidth}{%
    Analyze $S$ using $M \cup M_{org}$, and compute $l$ of $sw \in
    S$ using the forward-backward algorithm.
  }
  \label{line-analyze-corpus-end}
  \EndFor
  \label{line-update-likelihood}
  \State \parbox[t]{\dimexpr\linewidth-\leftmargin}{%
    Compute the new $l$ by summing all $(sw, l)$ in $C$, and update
    the likelihoods of $M$.
    \Comment{M step}
  }
  \State\Return $M$
  \EndFunction
\end{algorithmic} 
}
\end{algorithm}

Algorithm \ref{algo-additional-model} shows the learning algorithm of
the additional subword model.  This algorithm learns the additional
model $M$ that includes subwords that do not exist in the original
model $M_{org}$.

\begin{itemize}
\item
The \textsc{GenerateInitModel} function (Lines
\ref{line-init-model-start} to \ref{line-init-model-end}) generates the
initial model.  The initial model includes substrings that appear in 
corpus $C$ as subword candidates.

When we initialize the additional model from the training corpora, only
substrings that start with new letters, which are not included in the
original model, are added to the model as subword candidates.

\item
Lines \ref{line-main-loop-start} to \ref{line-main-loop-end} are the
main loop.  We update the likelihoods in model $M$ twice using the EM
algorithm (the \textsc{UpdateLikelihood} function).  When analyzing
the training corpora (Lines \ref{line-analyze-corpus-start} to
\ref{line-analyze-corpus-end}), we use a mixture of the original model
and additional model, but only update the additional model (Line
\ref{line-update-likelihood}).  Then, we eliminate 20 percent of the
low-likelihood subwords in one loop.
\end{itemize}

\section{Inuktitut Translation Experiments}
\label{sec-experiments}

In this paper, we evaluate our method for translation between English
(En) and Inuktitut (Iu), which was the shared task at the WMT-20
conference \cite{wmt-2020-machine}.  Table \ref{tbl-inuktutit-example}
shows an example of En-Iu translation.  Inuktitut is written using the
letters of the ``Unified Canadian Aboriginal Syllabics (U+1400 - U+167F)''
of the Unicode Standard.

Although the SentencePiece model of mBART-50 was trained on a corpus
that consisted of 100 languages and included 250K subwords, it does
not contain Unified Canadian Aboriginal Syllabics.  Therefore, if we
analyzed Inuktitut sentences using this model, subword segmentation
would fail, and most tokens would become out-of-vocabulary (OOV).%
\footnote{When we tokenize the Inuktitut language using the original
  tokenizer, sentences are segmented by spaces because SentencePiece
  bundles continuous unknown letters as a token.}

In the shared task at WMT-20, the training corpora were provided by
the organizers.  All participants trained their tokenizer from
scratch, and no teams used published multilingual pretrained models
\cite{zhang-etal-2020-niutrans,chen-etal-2020-facebook,kocmi-2020-cuni,roest-etal-2020-machine,knowles-etal-2020-nrc,bawden-etal-2020-university}.

\begin{table}[t]
\begin{center}
{\small
\begin{tabular}[t]{lp{50mm}}
\toprule
Language & Sentence \\
\midrule

English
& Good afternoon, Nunavummiut and residents of my community. \\
\midrule

\raisebox{1.0em}{Inuktitut}
& \includegraphics[clip,width=50mm]{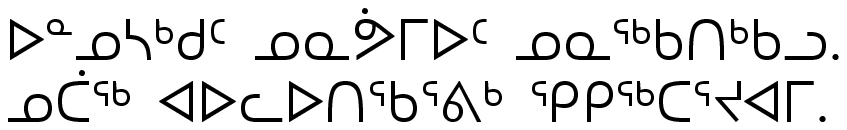} \\
\bottomrule
\end{tabular}
}
\caption{Example of English-Inuktitut translation.}
\label{tbl-inuktutit-example}
\end{center}
\end{table}

\subsection{Experimental Settings}

\begin{table}[t]
\begin{center}
{\small
\begin{tabular}{lrrr}
\toprule
Set & \#Sentences \\
\midrule

Training    & 1,308,277 \\
Development & 5,173 \\
Test        & 2,971 \\
\bottomrule
\end{tabular}
}
\caption{English-Inuktitut parallel corpus size.}
\label{tbl-inuktutit-corpora}
\end{center}
\end{table}

\paragraph{Corpora}

The size of the parallel corpus used in the shared task at WMT-20 is
shown in Table \ref{tbl-inuktutit-corpora}.

\paragraph{Tokenizer Setting}
We learned \{2K, 4K, 8K\} additional subwords from the Inuktitut side
of the parallel corpus.

We converted the unigram model provided from mBART-50 into a text
model using the \texttt{spm\_export\_vocab} command of SentencePiece.
We implemented the extension described in Sections \ref{sec-analysis},
\ref{sec-model-training}, and \ref{sec-subword-addition} using Python.
Note that we normalized the input sentences (including space
conversion) using the \texttt{spm\_normalize} command.

\paragraph{Baselines}
In our experiments, we set three baselines for tokenization.

\begin{itemize}\itemsep=0mm
\item \textbf{mBART-50 Tokenization Model}:
We used the SentencePiece model provided from mBART-50 with no change.

\item \textbf{Inuktitut Letters}:
After segmentation using the mBART-50 tokenization model,
we further divided the Inuktitut subwords into letters.

\item \textbf{Shared Vocabulary 32K Model}:
From the English-Inuktitut parallel corpus, we learned a shared model
using SentencePiece.  The vocabulary size was 32K.%
\footnote{We tested the vocabulary sizes of 16K, 32K, and 64K.  We
  employed the vocabulary size of 32K among them because because the
  BLEU score of the Transformer big model was the highest.}
\end{itemize}

\paragraph{Translation System and Model}

We used the Fairseq translator \cite{ott-etal-2019-fairseq}.

The mBART-50 model%
\footnote{\url{https://dl.fbaipublicfiles.com/fairseq/models/mbart50/mbart50.pretrained.tar.gz}}
that we used is an encoder-decoder model.  It can be used as a
multilingual translation model if we fine-tune it using parallel
corpora.  In this paper, we used the bidirectional model between
English and Inuktitut fine-tuned using the parallel corpus.

We used \newcite{wang-etal-2020-extending}'s approach to adapt
mBART-50 to a new language.  Specifically, we extended the word
embedding tables of the encoder and decoder as follows to achieve
Inuktitut translation using mBART-50:

\begin{itemize}\itemsep=0mm
\item
mBART-50 learns and translates sentences while appending language tags
in the source and target sequences.  To accept the new tags, we added
the Inuktitut language tag (\texttt{iu\_CA}) to the word embedding
tables.

\item
We also added the new subwords, which we had added to the tokenizer, 
to the word embedding tables.
\end{itemize}

\newcite{wang-etal-2020-extending} applied continued pretraining to
the extension of the word embedding tables using monolingual corpora.
By contrast, we randomly initialized the extention and learned it
during fine-tuning together with the other parameters.

As the baseline translator, we used a Transformer big model
\cite{DBLP:journals/corr/VaswaniSPUJGKP17}, which we trained from
scratch using only the parallel corpus tokenized by the shared
vocabulary 32K model.

\paragraph{Hyperparameters}
Table \ref{tbl-setting-traintest} shows the list of hyperparameters
during the fine-tuning of the mBART-50 model and testing.  Because the
total number of training tokens changes depending on the tokenizer, we
unified the warmup time to one epoch.

Note that we used the same hyperparameters for training the Transformer big
model and fine-tuning the mBART-50 model,
except for the learning rate and warmup time (LR=0.0004, warmup=5
epochs).

\begin{table}[t]
\begin{center}
{\small
\begin{tabular}{lp{50mm}}
\toprule
Training/Test & Hyperparameter \\
\midrule

Training
& LR=0.00008, dropout=0.3, \\
(Fine-tuning)
& the batch size was 8K tokens/update, \\
& warmup=around 1 epoch, \\
& LR\_scheduler=inverse\_sqrt, \\
& optimizer=adam, \\
& criterion=label\_smoothed\_cross\_entropy \\

\midrule
Test
& Beam width=10, \\
& length penalty=1.0 \\
\bottomrule
\end{tabular}
}
\caption{Hyperparameters for training and test.}
\label{tbl-setting-traintest}
\end{center}
\end{table}

\subsection{Results of Tokenization}

\begin{table*}
\begin{center}
{\small
\begin{tabular}{llrrr}
\toprule
& & & 
& Additional Words \\

Type
& Tokenizer
& \#Tokens/Sent.
& OOV Rate 
& for No OOVs \\

& 
& in Test Set
& in Test Set
& in Training Set \\
\midrule

Baselines
& mBART-50 Tokenization Model
& 20.9
& 43.1\% 
& 1,553,466 \\

& Inuktitut Letters
& 80.0
& 85.1\% 
& 141 \\

& Shared Vocabulary 32K Model
& 22.0
& 82.7\% 
& 26,657 \\
\midrule

Our Method
& 2K Additional Subwords
& 37.9
& 68.7\% 
& 2,001 \\

& 4K Additional Subwords
& 34.6
& 65.7\%
& 4,001 \\

& 8K Additional Subwords
& 31.8
& 62.8\% 
& 8,001 \\
\bottomrule
\end{tabular}
}
\caption{Number of tokens for a sentence in the Inuktitut test
  set, OOV rate viewed from the original vocabulary of mBART-50, and
  number of additional words to eliminate OOVs.}
\label{tbl-sentence-length}
\end{center}
\end{table*}

\begin{table*}
\begin{center}
{\small
\begin{tabular}{llrrr@{\hspace*{0.5em}}lr@{\hspace*{0.5em}}l}
\toprule
& & &
& \multicolumn{4}{c}{BLEU} \\

Translation Model
& Tokenizer
& \#Added Emb.
& OOV Rate
& \multicolumn{2}{c}{En $\rightarrow$ Iu}
& \multicolumn{2}{c}{Iu $\rightarrow$ En} \\
\midrule

Transformer Big Model
& Shared Vocabulary 32K Model
& --- & 0.0\%
& 8.3 & & 21.5 & \\
\midrule

mBART-50 Model
& mBART-50 Tokenization Model
& 0 & 43.1\%
& 1.7 & (-) & 7.0 & (-) \\

& & 1,553,466 & 12.1\%
& N/A & & N/A & \\

& Inuktitut Letters
& 140 & 0.0\%
& 9.8 & (+) & \textbf{23.5} & (+) \\

& Shared Vocabulary 32K Model
& 26,657 & 0.0\%
& 9.6 & (+) & 22.8 & (+) \\
\cmidrule{2-8}

& 2K Additional Subwords
& 2,000 & 0.0\%
& \textbf{10.2} & (+\dag) & \textbf{23.5} & (+) \\

& 4K Additional Subwords
& 4,000 & 0.0\%
& 9.7 & (+) & 23.3 & (+) \\

& 8K Additional Subwords
& 8,000 & 0.0\%
& 9.8 & (+) & 23.2 & (+) \\
\bottomrule
\end{tabular}
}
\caption{BLEU scores for English and Inuktitut translation.  Bold
  values indicate the highest BLEU scores in each direction.  (+) and
  (-) marks indicate that the score was significantly better and worse
  than that of the Transformer big model (with shared vocabulary 32K
  tokenization), respectively.  (\dag) indicates that the score was
  significantly better than that of the mBART-50 model (Inuktitut
  letter tokenization).}
\label{tbl-inuktitut-translation}
\end{center}
\end{table*}

Table \ref{tbl-sentence-length} shows the results of the tokenization
of Inuktitut sentences in the test set.  It shows the number of tokens
for a sentence and the OOV rate, which we measured using the
vocabulary of the translation model.
Although all tokenizers had high OOV rates, we could reduce them by
adding new words (embeddings) to the vocabulary of mBART-50.  The
``additional words for no OOVs'' in Table \ref{tbl-sentence-length}
indicate the number of words to add to the vocabulary.

In neural machine translation, it is advantageous if the number of
tokens for a sentence is small.  In our experiment, the least number
of tokens was obtained by the tokenization using the mBART-50
tokenization model.  This was because SentencePiece tokenized the
sentences using space characters, and over 40\% of tokens became OOV.
It is quite difficult to train a translation model using this
tokenizer because we must add over 1.5 million words into the
vocabulary of mBART-50.%
\footnote{The vocabulary size of the Inuktitut language tends to be
  large because it has inflection and agglutination.}

The shared vocabulary 32K model achieved the next smallest number of
tokens.  However, we would have to add 27K out of 32K subwords because
the vocabulary of this model is different from that of mBART-50.

In the case of the Inuktitut letters, although we would have to add
only 141 subwords to the vocabulary to achieve no OOVs, the length of
input/output sequences became long, that is, 80 tokens per sentence.

By contrast, the numbers of tokens obtained by our method were less
than half that of the ``Inuktitut letters'' case, even though they
were larger than those of the mBART-50 tokenization model and shared
vocabulary 32K model.%
\footnote{The number of additional words required to maintain no OOVs
  and that for the tokenizer did not match.  This is because the
  tokenizer was trained using the monolingual corpus, and the
  vocabulary contained subwords that did not appear in the parallel
  corpus.}
Our tokenizer can eliminate OOVs while controlling the
number of additional subwords.

When we tokenized the English sentences in the test set using our
tokenizers, we obtained two sentences with different tokenizations.
The differences were that continuous periods `....' were segmented
into `.$\sqcup$...' or `...$\sqcup$.', which rarely affected the
translation.

\subsection{Translation Results}

The translation results are shown in Table
\ref{tbl-inuktitut-translation}.  We used a Transformer big model
\cite{DBLP:journals/corr/VaswaniSPUJGKP17} as the baseline.  We
fine-tuned the mBART-50 translation models after we extended the word
embeddings of ``\#Added Emb.''  ``OOV Rate'' indicates the OOV rate in
the test set after we extended the word embeddings.
For the evaluation, we used sacreBLEU \cite{post-2018-call} and
performed a significance test using bootstrap resampling, implemented in
sacreBLEU, with 5\% of the significance rate ($p < 0.05$).

First, comparing the translation models, the BLEU scores of the
mBART-50 pretrained model improved from that of the Transformer
big model in both the En-Iu and Iu-En directions, regardless of the
tokenizer (except for the mBART-50 tokenization model).  Even though
the mBART-50 model does not include Inuktitut, pretraining was
effective in improving the BLEU scores.

Focusing on the tokenizers in the mBART-50 translation model, we could
not obtain a meaningful translation using the original mBART-50
tokenization model because of the high OOV rate.  To reduce the OOV rate,
we added over 1.5 million words, but we could not fine-tune the
mBART-50 pretrained model because of the memory limitation.
Among the baseline tokenizers, Inuktitut letter tokenization had the
highest BLEU score.  The BLEU score of the shared vocabulary 32K model
was lower than that of the Inuktitut letters.

Compared with Inuktitut letter tokenization, the BLEU score of the
``2K Additional Subwords'' of En-Iu translation was significantly
higher, but significant differences were not observed in the other
tokenizers of ours.  The small number of the additional subwords tend
to be high BLEU score.  We suppose these results indicate that
fine-tuning alleviated the difference between the tokenizers.

To summarize, 
we could apply the multilingual pretrained model mBART-50
to Inuktitut translation by segmenting Inuktitut sentences into subwords,
and consequently improved the BLEU scores.

\section{Conclusions}
\label{sec-conclusions}

In this paper, we added new subwords to the SentencePiece tokenizer to
apply a multilingual pretrained model to new languages.  The proposed
method added new subwords with unknown letters and their likelihood to
the original model without changing previous tokenization results.

In our experiments, we segmented Inuktitut sentences into subwords,
while controlling the number of additional subwords.  We applied our
tokenizer to the mBART-50 pretrained model for Inuktitut translation.
As a result, the BLEU scores improved for the new language.

Although we estimated only additional subwords in this study, our method
can re-estimate the likelihood of existing subwords.  In future
work, we will evaluate the effectiveness of adding arbitrary subwords to
the original models because some researchers have reported that
optimized tokenization improves the accuracy of downstream tasks
\cite{he-etal-2020-dynamic,hiraoka-etal-2021-joint}.

We release our implementation for learning additional subword models
via GitHub.%
\footnote{\url{https://github.com/kenji-imamura/sentpiece_mimic}}

\section*{Acknowledgments}

Part of this work was conducted under the commissioned research
program `Research and Development of Advanced Multilingual Translation
Technology' in the `R\&D Project for Information and Communications
Technology (JPMI00316)' of the Ministry of Internal Affairs and
Communications (MIC), Japan.

\bibliography{anthology,custom}
\bibliographystyle{acl_natbib}

\end{document}